\documentclass[10pt,twocolumn,letterpaper]{article}

\usepackage[pagenumbers]{cvpr} 

\usepackage[dvipsnames]{xcolor}

\usepackage{array}
\usepackage{color}
\usepackage{epsfig}
\usepackage{graphicx}
\usepackage{textcomp}
\usepackage{gensymb}
\usepackage{booktabs}
\usepackage{makecell}
\usepackage{enumitem}
\usepackage[ruled,vlined]{algorithm2e}
\usepackage[normalem]{ulem}
\usepackage{multirow}
\usepackage{wrapfig}
\usepackage{xspace}
\usepackage[accsupp]{axessibility} 

\newcommand\boldblack[1]{\textcolor{black}{\textbf{#1}}}
\newcommand{\marepo}[0]{\textbf{\textit{marepo}}\xspace}

\definecolor{cvprblue}{rgb}{0.21,0.49,0.74}
\usepackage[pagebackref,breaklinks,colorlinks,citecolor=cvprblue]{hyperref}

\title{Map-Relative Pose Regression for Visual Re-Localization}

\author{Shuai Chen$^{1,2}$ \qquad Tommaso Cavallari$^1$ \qquad Victor Adrian Prisacariu$^{1,2}$ \qquad Eric Brachmann$^1$ \\
$^1$Niantic \qquad $^2$University of Oxford
}

\begin{document}

\maketitle
\begin{abstract}
Pose regression networks predict the camera pose of a query image relative to a known environment.
Within this family of methods, absolute pose regression (APR) has recently shown promising accuracy in the range of a few centimeters in position error.
APR networks encode the scene geometry implicitly in their weights.
To achieve high accuracy, they require vast amounts of training data that, realistically, can only be created using novel view synthesis in a days-long process.
This process has to be repeated for each new scene again and again.
We present a new approach to pose regression, map-relative pose regression (\marepo), that satisfies the data hunger of the pose regression network in a scene-agnostic fashion. 
We condition the pose regressor on a scene-specific map representation such that its pose predictions are relative to the scene map. 
This allows us to train the pose regressor across hundreds of scenes to learn the generic relation between a scene-specific map representation and the camera pose.
Our map-relative pose regressor can be applied to new map representations immediately or after mere minutes of fine-tuning for the highest accuracy.
Our approach outperforms previous pose regression methods by far on two public datasets, indoor and outdoor.
Code is available: \url{https://nianticlabs.github.io/marepo}.
\end{abstract}
\section{Introduction}
\label{sec:intro}
Today, neural networks have conquered virtually all sectors of computer vision, but there is still at least one task that they struggle with: visual relocalization.
What is visual relocalization? Given a set of mapping images and their poses, expressed in a common coordinate system, build a scene representation.
Later, given a query image, estimate its pose, i.e. position and orientation, relative to the scene.

Successful approaches to visual relocalization rely on predicting image-to-scene correspondences, either via matching~\cite{Sattler17,sarlin2019HFNet,compression2019cvpr,Sarlin20,kapture2020,sarlin21pixloc,zhou2022gomatch} or direct regression~\cite{Brachmann17,yang2019sanet,brachmann2020dsacstar,brachmann2023ACE,dong2022visual}, then solving for the pose using traditional and robust algorithms like PnP~\cite{gao2003complete} and RANSAC~\cite{Fischler81}.

Adopting a different perspective, approaches based on pose regression~\cite{Kendall15,Shavit21,chen2022dfnet,Moreau21} attempt to perform visual relocalization without resorting to traditional pose solving, by using a single feed-forward neural network to infer poses from single images.
The mapping data is treated as a training set where the camera extrinsics serve as supervision.
Generally, pose regression approaches come in two flavors, but they both struggle with accuracy compared to correspondence-based methods. 

Absolute pose regression (APR) methods ~\cite{Kendall15,Kendall17,Brahmbhatt18} involve training a dedicated pose regressor for each individual scene, enabling the prediction of camera poses to that particular scene.
Though the scene coordinate space can be implicitly encoded in the weights of the neural networks, absolute pose regressors exhibit low pose estimation accuracy, primarily due to the often limited training data available for each scene, and struggle to generalize to unseen views~\cite{Sattler19}.

\begin{figure}[t]
    \centering
    \footnotesize
    \includegraphics[width=\linewidth]{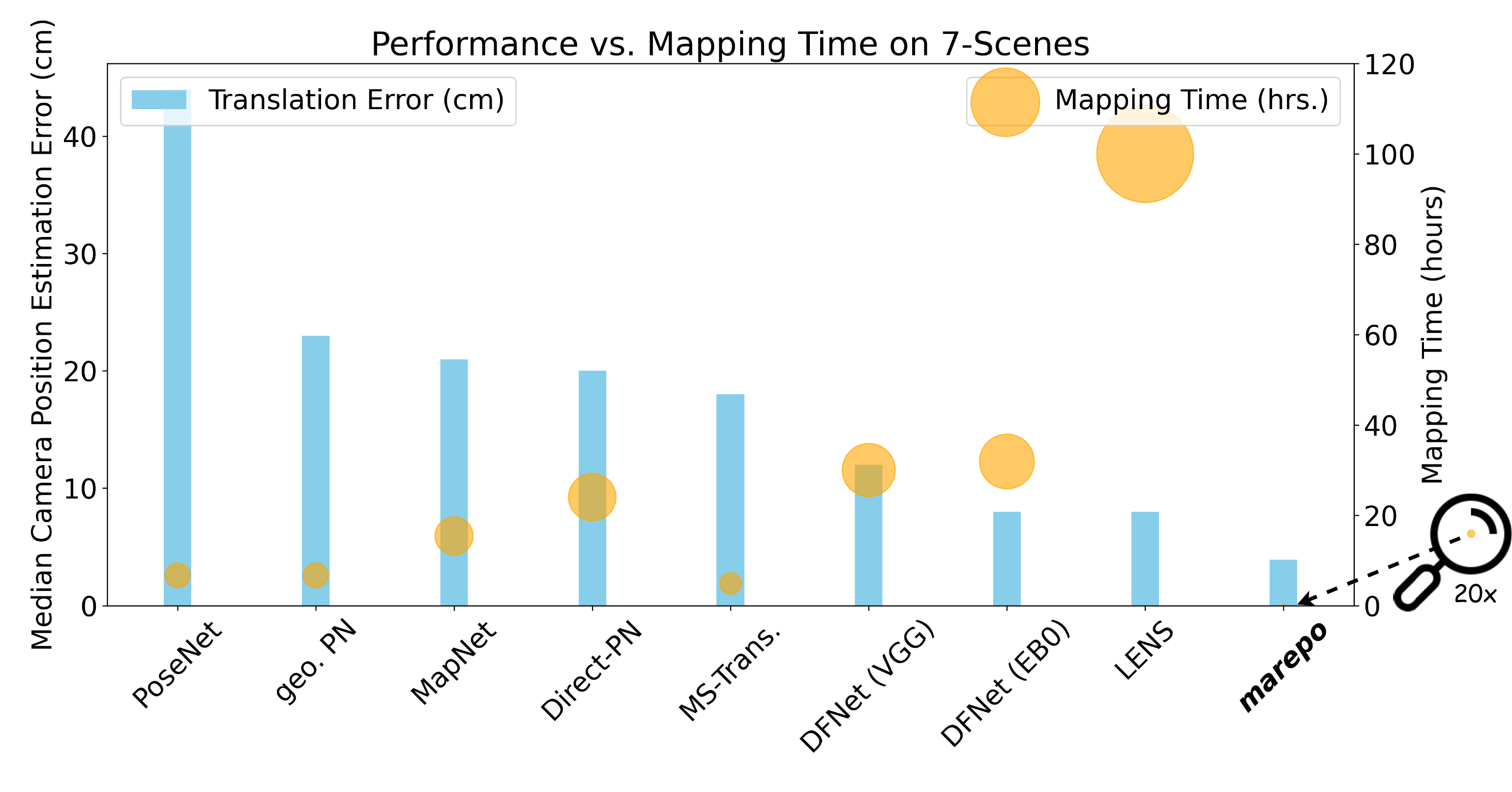}
    \caption{
    \textbf{Camera pose estimation performance vs. mapping time.}
    The figure shows the median translation error of several pose regression relocalization methods on the 7-Scenes dataset and the time required (proportional to the bubble size) to train each relocalizer on the target scenes.
    Our proposed approach, \marepo, achieves superior performance -- by far -- on both metrics, thanks to its integration of scene-specific geometric map priors within an accurate, map-relative, pose regression framework.
    }
    \label{fig:teaser}
\end{figure}

Relative pose regression is a second flavor of pose regression methods~\cite{turkoglu2021visual,WinkelbauerICRA21,en2018rpnet,chen2021wide,levinson2020analysis}.
The regressor is trained to predict the relative pose between two images.
In a typical inference scenario, the regressor is applied to a pair formed by an unseen query and an image from the mapping set (typically selected via a nearest neighbor-type matching); then, the predicted relative pose can be combined with the known pose of the mapping image to yield the absolute query pose. 
These methods can be trained on a lot of scene-agnostic data, but their accuracy is still limited: a metric pose between two images can only be predicted approximately~\cite{arnold2022mapfree}.

Motivated by those limitations, we propose a new flavor of absolute pose regression: map-relative pose regression (\marepo).
We couple a scene-specific representation -- encoding the scale-metric reference space of each target scene -- with a general, scene-agnostic, absolute pose regression network.
In particular, we utilize a fast-training scene coordinate regression model as our scene representation and train, once and ahead of time, a pose regression network that learns the relationship between a scene coordinate prediction and the corresponding camera pose.
This generic relationship allows us to train the pose regressor on hundreds of different scenes, effectively solving the issue of the limited availability of training data afflicting absolute pose regression models.
On the other hand, since at localization time our pose regressor is conditioned on a scene-specific map representation, it is able to predict accurate scale-metric poses, unlike relative pose regressors.

Our experiments show that \marepo is a pose regression network with an accuracy on par with structure-based relocalization methods (e.g.~\cite{brachmann2023ACE}), exceeding the accuracy of all other single-frame absolute pose regression methods by far (see~\cref{fig:teaser}).
Our scene-agnostic pose regressor can be applied to each new scene representation right away, or (optionally) fine-tuned in just a few minutes for best accuracy. 

\noindent We summarize our main contributions as follows:
\begin{enumerate}[leftmargin=*]
    \item We propose \marepo, a novel Absolute Pose Regression approach that combines a generic and scene-agnostic \underline{Ma}p-\underline{Re}lative \underline{Po}se regression method with a scene-specific metric representation. We show that the network can perform end-to-end inference on previously unseen images and, thanks to the strong and explicit 3D geometric knowledge encoded by the scene-specific component, it can directly estimate accurate, absolute, metric poses.
    
    \item We introduce a transformer-based network architecture that can process a dense set of correspondences between 2D locations in a query image and their corresponding 3D coordinates within the reference system of a previously mapped scene, and estimate the pose of the camera that captured the query image.
    We further show how a dynamic position encoding applied to the 2D locations in the query image can significantly improve the performance of the method by encoding the intrinsic camera parameters within the transformer input.
\end{enumerate}

\section{Related Works}
\label{sec:related_work}
Over the years many efforts in the literature have tackled the problem of visual relocalization, and we have seen a rough demarcation of the types of approach into two main fields: the more traditional approaches, relying on geometric concepts and the estimation of correspondences between images and maps; and the more recent ``direct'' approaches, relying on neural networks to predict the absolute position and orientation of the image without an intermediate, explicit, matching step linking the 2D image realm with a 3D map of the scene.
In the remainder of this section, we briefly explore the main approaches in each of these categories.

\subsection{Geometry-based Visual Relocalization}
Geometry-based approaches rely on estimating correspondences between pixels in the query images and points in the scene's map.
These correspondences effectively establish a 2D-to-3D matching that can be exploited by pose-solving methods such as PnP/RANSAC~\cite{gao2003complete,Fischler81} to compute the pose of the camera at the moment it captures the image.

There are several ways to estimate those correspondences: from classic computer vision approaches using off-the-shelf feature detectors and descriptors to compute matches between image pixels and a database of previously observed 3D points~\cite{Sattler12,Sattler17,compression2019cvpr,kapture2020}; to more advanced, neural-based approaches that rely on learned descriptors, improved matchers, and different map representations in order to estimate better correspondences from more challenging images or viewpoints~\cite{sarlin2019HFNet,Lynen2019largescale,sarlin21pixloc,panek2022meshloc}.
These approaches leverage the underlying geometric principles governing image formation and capture, yielding accurate pose estimations with low errors, often in the order of a few centimeters.
However, they are not without a drawback: they generally require the creation of a map (e.g., in the form of a 3D point cloud created via Structure-from-Motion) of the scene ahead of time in order to associate the descriptors to 3D coordinates, and that is typically time-consuming.

In recent years, a new approach to geometry-based relocalization started to become prominent: scene coordinate regression (SCR).
In this scenario, the map of the scene is directly encoded in a fixed-size set of weights of a neural network.
At localization time, the query image is passed through the network, yielding per-pixel scene coordinates that can be directly used by a pose solver to estimate the camera pose~\cite{Brachmann17,brachmann2019ngransac,yang2019sanet,brachmann2020dsacstar,li2020hierarchical,dong2022visual,brachmann2023ACE}. 
While effective, these methods have typically required training a new network for every new target scene, potentially taking several hours~\cite{brachmann2020dsacstar}, thus hindering their large scale application.
Recently, an approach to scene coordinate regression that can take mere minutes to be trained for every scene was presented in ~\cite{brachmann2023ACE}, making practical deployment of SCR networks a possibility.
As the correspondence-based methods mentioned above, coordinate regression approaches are also very accurate by relying on geometric information on the structure of the scene.
Nevertheless, scene coordinate regression methods still require an explicit stage where a pose solver has to process each correspondence generated by the method to estimate the camera pose.
Conversely, Absolute Pose Regression methods do not have this requirement since the regressor network can go directly from image to pose in an end-to-end fashion.

\subsection{Absolute Pose Regression}
Absolute Pose Regression (APR) approaches have also garnered notable attention recently, primarily due to their simplicity and efficiency.
These methods directly predict camera poses via end-to-end neural networks. Kendall et al. introduced the first APR approach, named PoseNet~\cite{Kendall15, Kendall16, Kendall17},
where a feed-forward neural network directly regresses a 7-dimensional pose vector for every query image.
Successive works explore diverse architectural designs such as hourglass networks \cite{Melekhov17}, bifurcated translation and rotation regression \cite{Wu17,Naseer17}, attention layers \cite{atloc,Shavit21multiscene,Shavit21,shavit2022camera}, and LSTM layers \cite{Walch17}.
Other research efforts attempt to improve APR performance with different supervisions, such as a geometric loss \cite{Kendall17}, relative pose constraints \cite{Brahmbhatt18}, uncertainty awareness \cite{Kendall17,moreau2022coordinet}, or a sequential formulation like temporal filtering \cite{clark2017vidloc} and multitasking \cite{Radwan18}.
Despite these advancements, the accuracy of single-frame-based pose regression remains limited when compared to alternative approaches, such as those based on geometric principles.

Among recent advancements within APR, a promising direction is incorporating novel view synthesis techniques, either by synthesizing large amounts of training data to solve overfitting issues~\cite{Purkait18, Moreau21, chen2022dfnet} or by integrating them into a fine-tuning process before test time~\cite{Brahmbhatt18, chen21, chen2022dfnet}.
One drawback of the former is that generating high-quality synthetic data can be a time-intensive process; as for the latter, in addition to time requirements, those approaches typically require extra data from the scene of interest.
These limitations pose significant constraints in environments subject to rapid changes, such as those with frequent alterations in furnishings or appearances.

This paper introduces a new category of approach to the pose regression domain, one that tops the need for extensive mapping time, reducing it to mere minutes per scene.
It demonstrates enhanced accuracy over previous single-frame APR methods and exhibits rapid scalability to new environments, making it flexible to deploy in a fast-changing world as we live in today.

\section{Method}
\label{sec:method}
\begin{figure*}[t]
    \centering
    \footnotesize
    \includegraphics[width=0.8\linewidth]{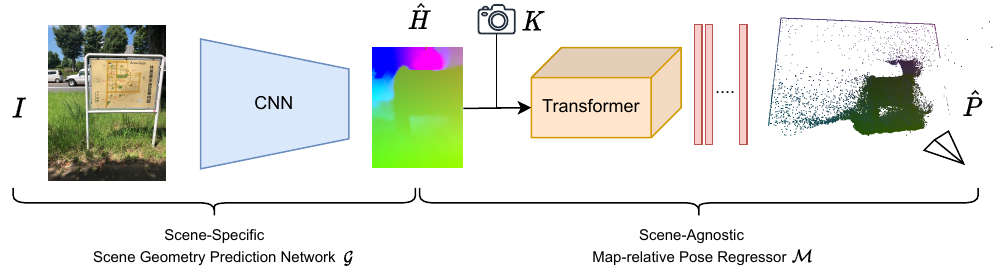}
    \caption{
    Illustration of the \marepo network.
    A scene-specific geometry prediction module $\mathcal{G_S}$ processes a query image to predict a scene coordinate map $\hat{H}$.
    Then, a scene-agnostic map-relative pose regressor $\mathcal{M}$ is used to directly regress the camera pose.
    Our network's training and inference rely solely on RGB images $I$ and camera intrinsics $K$ without requiring depth information or pre-built point clouds.
    }
    \label{fig:ACEFormer}
\end{figure*}

Prevalent pose regression methods are built on top of end-to-end neural network-based approaches.
They can be formulated as $\hat{P} = \mathcal{F}(I)$: the camera poses $\hat{P}$ are directly predicted by providing an input image $I$ to the network $\mathcal{F}$.
A benefit of this type of approach is its conceptual simplicity and highly efficient inference speed.
However, the forward process of typical APR networks -- which rely on 2D operations over images and features -- does not exploit any 3D geometric reasoning, resulting in insufficient performance compared to state-of-the-art geometry-based methods.
In this paper, we propose a first map-relative pose regression approach empowered with explicit 3D geometric reasoning within its formulation, allowing us to regress accurate camera poses while maintaining real-time efficiency and end-to-end simplicity like any other pose regression method.

In the remainder of this section we first give an overview of the transformer-based network architecture we deploy to perform pose regression (\cref{sec:arch-overview}); then we describe the main components and ideas behind the proposed approach (\cref{sec:arch-details}); the loss function optimized during training (\cref{sec:arch-loss}); and, finally, we show how the scene-agnostic pose-regression transformer can be \textit{optionally} fine-tuned for a specific testing scene in a matter of minutes, thus improving the performance of the method even further compared to a non-fine-tuned regressor
(\cref{sec:arch-finetuning}).

\subsection{Architecture Overview}
\label{sec:arch-overview}
The main architecture of our method is formed of two components: (1) a CNN-based scene geometry prediction network $\mathcal{G}$ that maps pixels from the input image to 3D scene coordinates; and (2) a transformer-based map-relative pose regressor $\mathcal{M}$ that, given the scene coordinates, estimates the camera poses.
Ideally, the network $\mathcal{G}$ is designed to associate each input image to scene-specific 3D information, thus requiring some training process for every new scene processed by the method.
Conversely, the map-relative pose regressor $\mathcal{M}$ is a scene-agnostic module trained with large amounts of data and can generalize to unseen maps.

We illustrate our proposed network architecture in \cref{fig:ACEFormer}.
Given an image $I$ from scene $S$, we pass it to our model which outputs a pose $\hat{P}$.
The process is formulated as:
\begin{equation}
    \hat{P} = \mathcal{M}(\hat{H}, K) = \mathcal{M}(\mathcal{G_S}(I), K),
\end{equation}
where $\hat{H} = \mathcal{G_S}(I)$ indicates the image-to-scene coordinates predicted by $\mathcal{G}$ (which was trained ad-hoc for scene $S$), and $K \in\mathbb{R}^{3 \times 3}$ is the camera intrinsic matrix associated with the input image.
This formulation makes the approach similar to both standard Absolute Pose Regression, in that it generates poses via a feed-forward pass through a neural network, as well as Scene Coordinate Regression since the scene geometry prediction network regresses 3D coordinates directly from each input image.
Unlike standard APR, our method has full geometric reasoning on the link between the image and the scene, and, unlike SCR approaches, it does not require a traditional, non-deterministic RANSAC stage to infer the pose.
Theoretically, any algorithm capable of predicting 3D scene coordinates from an input image could be a viable candidate as $\mathcal{G}$, since the following transformer we deploy to perform pose regression ($\mathcal{M}$) does not depend on the prior component.

In the next section, we will focus on detailing the map-relative pose regression network $\mathcal{M}$.

\subsection{Map-Relative Pose Regression Architecture}
\label{sec:arch-details}
\begin{figure}[t]
    \centering
    \footnotesize
    \includegraphics[width=0.8\linewidth]{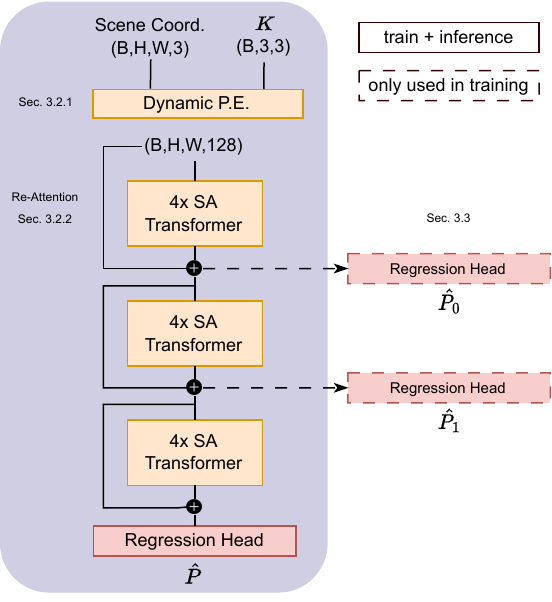}
    \caption{
    The map-relative pose regressor $\mathcal{M}$
    takes as input a tensor of predicted scene coordinate maps and the corresponding camera intrinsics, 
    embeds the information with dynamic positional encoding into higher dimensional features, and finally estimates the camera poses $\hat{P}$.
    During training, we also predict $\hat{P_0}$ and $\hat{P_1}$ for intermediate supervision.
    }
    \label{fig:transformer}
    \vspace{-2em}
\end{figure}
To achieve a robust and scene-agnostic map-relative pose regression, we carefully design the simple yet effective architecture depicted in \cref{fig:transformer}.
The main components of this module are: (a) a novel dynamic positional encoding used to increase the dimensionality of the input scene coordinates -- as well as embed their spatial location within the input image -- taking into account the intrinsic properties of the camera that captured the frame; (b) several multi-layer self-attention transformer blocks; and finally (c), an MLP-based pose regression head.
Given scene coordinate maps $\hat{H}$ (predicted by $\mathcal{G_S}$) and the corresponding camera intrinsic matrices $K$, the network is able to directly estimate 6-DoF (six degrees of freedom) metric camera poses.
We detail the designs of each component in the following sub-sections.

\subsubsection{Dynamic Positional Encoding}
Unlike many vision transformers (ViTs) for high-level tasks~\cite{carion2020DeTR, dosovitskiy2020image, sun2021loftr}, where the transformer is conditioned to operate directly upon input RGB images (or higher-dimensional features), our transformer is designed to interpret accurate 3D geometric information strongly connected to real-world physics.
The content a camera captures in its frame is strictly associated with its intrinsic parameters; thus, we propose to use a positional encoding that is conditioned on each individual sensor, allowing us to train the main transformer blocks in a fashion that is generic, i.e., independent of the camera calibration parameters.

Our positional encoding scheme entails the fusion of two different components: (1) a camera-aware 2D positional embedding, associating each predicted scene coordinate to its corresponding pixel location;
and (2) a 3D positional embedding that embeds the actual 3D scene coordinate values into a high-frequency domain.

\paragraph{Camera-Aware 2D Positional Embedding}
We draw inspiration from LoFTR's~\cite{sun2021loftr} positional embedding, but integrate information from the camera's intrinsics to generate the high-frequency components that are fed to the network.

Specifically, for each pixel coordinate $(u, v)$ in the input image, we first compute the $(x, y)$ components of the 3D ray originating in the camera center and passing through the pixel (ignoring the $z$ component);
then apply the positional embedding from \cite{sun2021loftr} on the (now camera-invariant) ray's directional components.
This generates a high-frequency/high-dimensionality embedding, allowing the transformer to correlate the input 3D coordinates (predicted by $\mathcal{G_S}$ and defined in a scene-specific coordinate system) with 3D rays originating from the current camera position, helping with the task of regressing the current camera pose w.r.t. the origin of the scene coordinate system.
Formally, we define the Camera-Aware 2D Positional Embedding as follows:
\begin{equation}
\mathcal{P} \mathcal{E}_{2D\left(u, v\right)}^i:=\left\{\begin{array}{ll}
\sin \left(\omega_k \cdot X_{\text{ray}}(u)\right), & i=4 k \\
\cos \left(\omega_k \cdot X_{\text{ray}}(u)\right), & i=4 k+1 \\
\sin \left(\omega_k \cdot Y_{\text{ray}}(v)\right), & i=4 k+2 \\
\cos \left(\omega_k \cdot Y_{\text{ray}}(v)\right), & i=4 k+3
\end{array},\right.
\end{equation}
where $\omega_k=\frac{1}{10000^{2 k / d}}$ is the frequency band defined for $d$-dimensional features in which positional encoding is applied on, $i$ is the current feature index, and $X_{\text{ray}}$ and $Y_{\text{ray}}$ are the $X$ and $Y$ components of the rays passing through $(u, v)$:
\begin{equation}
\begin{aligned}
X_{\text{ray}}(u) & =\lambda\frac{u - c_x-\varepsilon}{f_x} \text {,} \\
Y_{\text{ray}}(v) & =\lambda\frac{v - c_y-\varepsilon}{f_y} \text {,} \\
\end{aligned}
\end{equation}
with $f_{x/y}$ and $c_{x/y}$ corresponding to the intrinsics of the input frame, $\varepsilon=0.5$ to achieve zero-mean in the center of the image, and $\lambda=400$ chosen as a heuristic constant to keep a reasonable numerical magnitude for the final embedding.

\paragraph{3D Positional Embedding}
We use a 3D positional embedding to map the scene coordinates $p \in \mathbb{R}^3$ predicted by $\mathcal{G_S}$ to high frequency/dimensionality, inspired by \cite{Mildenhall20nerf}:
\begin{equation}
\begin{aligned}
& \mathcal{P}\mathcal{E}_{3D}(p)=\operatorname{Conv}_{3(2m+1)}^{d} [p, \sin\left(2^{0}\pi p\right), \cos\left(2^{0}\pi p\right),\\
& \quad \quad \quad \quad \quad \ldots, \sin \left(2^{m-1} \pi p\right), \cos \left(2^{m-1} \pi p\right)].
\end{aligned}
\end{equation}
Here, in addition to the sinusoidal embedding mapping the 3D coordinates to a $3(2m + 1)$-dimensional space, we also apply a further $1 \times 1$ convolution $\operatorname{Conv}_{6m+3}^{d}$ to ensure both $\mathcal{P} \mathcal{E}_{2D}$ and $\mathcal{P} \mathcal{E}_{3D}$ have the same number of channels.

\paragraph{Fused Positional Embedding}
Finally, we fuse the 2D and 3D embeddings before passing them to the transformer:
\begin{equation}
\mathcal{P}\mathcal{E}_{f} = \mathcal{P}\mathcal{E}_{3D} + \mathcal{P}\mathcal{E}_{2D}.
\end{equation}

\subsubsection{Re-Attention for Deep Transformer}
As illustrated in the left part of \cref{fig:transformer}, the core of our map-relative pose regression architecture is formed by twelve self-attention transformers arranged over three blocks of four transformers each.
In our implementation, we use Linear Transformers~\cite{katharopoulos2020transformers} as they reduce the computation complexity of each layer from quadratic to linear in the length of the input (i.e., the resolution of the scene coordinate map).

Since the Dynamic Positional Encoding is fed to the network only at the beginning, we found that the information flow became weaker as the depth of the network increases.
To solve this problem, we add what we call a ``Re-Attention'' mechanism, introducing residual connections every four blocks.
Experimentally, we find that this practice is quite effective, allowing the network to converge more quickly and leading to a better generalization.

\subsubsection{Pose Regression Head}
The last component of the \marepo architecture is a pose regression head.
Its structure is simple: first, a residual block formed of three $1 \times 1$ convolution layers followed by global average pooling generates a single embedding that represents the whole input scene-coordinate map.
Such embedding is then passed to a small MLP (3 layers) that directly outputs the camera pose as a 10-dimensional representation.
The pose representation can then be unpacked into translation and rotation: the translation is represented by four homogeneous coordinates (inspired by~\cite{brachmann2023ACE}); the rotation is encoded as a 6D vector representing two un-normalized axes of the coordinate system that are later used to form a full rotation matrix by normalization and cross-product, as in~\cite{Zhou_2019_CVPR}.

\subsection{Loss Function}
\label{sec:arch-loss}
The map-relative pose regressor architecture described above is able to directly output a metric pose $\mathcal{\hat{P}}$ (formed of a $3 \times 3$ rotation matrix $\hat{R}$, and a translation vector $\mathbf{\hat{t}}$) for each image.
In order to train such system we use a standard L1 pose regression loss proposed in~\cite{arnold2022mapfree}, defined as follows:
\begin{equation}
\mathcal{L}_{\hat{P}}=\|\hat{R}-R\|_1 + \|\hat{\mathbf{t}}-\mathbf{t}\|_1.
\end{equation}
Experimentally, we found that adding supervision at intermediate layers of the regressor is beneficial to the overall performance.
Therefore, at training time, we additionally apply the pose regression head after each block of four self-attention transformers (see~\cref{fig:transformer}, right) and compute auxiliary losses $\mathcal{L}_{P_0}$ and $\mathcal{L}_{P_1}$ as described above.
Thus, the overall loss we optimize during training is as follows:
\begin{equation}
    \mathcal{L} = \mathcal{L}_{\hat{P_0}} + \mathcal{L}_{\hat{P_1}} + \mathcal{L}_{\hat{P}}, 
    \label{eq:Loss}
\end{equation}
During inference we only use the last output pose, $\hat{P}$.

\subsection{(Optionally) Fine-Tuning the Pose Regressor}
\label{sec:arch-finetuning}
As described earlier, the proposed map-relative pose regressor is formed by two main components: an initial scene-specific network $\mathcal{G_S}$ that is able to predict metric scene coordinates for each pixel (in our implementation, we use an off-the-shelf scene coordinate regression architecture that can be trained in a few minutes for each scene~\cite{brachmann2023ACE}); 
and a scene-agnostic regressor $\mathcal{M}$ that exploits the geometric information encoded by the scene coordinates to predict the camera pose.
The latter is trained once -- ahead of time -- over a large corpus of data, but it is reusable \textit{as-is} for every new target scene.
We find that this hybrid approach works exceptionally well compared to other APR methods that are trained with the traditional end-to-end image-to-pose protocol, over the course of hours or days.

Still, we also explore whether applying a scene-specific adaptation stage to the transformer-based regressor can be beneficial to the performance of the method.
In this scheme, for each new scene being evaluated, after training the scene-specific coordinate regressor $\mathcal{G_S}$, we fine-tune the pose-regressor $\mathcal{M}$ on the same mapping images, using the same loss as in \cref{sec:arch-loss}.
Fine-tuning the transformer is very efficient in terms of resources required: in the next section we show that, with only two passes over the training dataset for each new scene (taking typically between 1-10 minutes, depending on the number of frames), our method can further improve its performance from what was already state-of-the-art compared to pose regression-based methods.

Notably, the final fine-tuning step is completely \textit{optional} in our approach: the pre-trained $\mathcal{M}$ is already capable of predicting accurate camera poses, given 3D scene coordinates predicted by the geometric network module $\mathcal{G_S}$.

\section{Experiments}
\label{sec:exp}

\begin{table*}[t]
\footnotesize
\caption{\textbf{Re-localization results on the indoor 7-Scenes dataset.}
Pose errors are shown as median translation (cm) and rotation (\degree) errors.
Numbers in \boldblack{bold} represent the best performance among the APR-based approaches.
$\marepo$ denotes our model with generic transformer-based pose regressor $\mathcal{M}$.
$\marepo_{S}$ reports the performance of the model after $\mathcal{M}$ has been fine-tuned for each scene.
}
\label{table:1}
\vspace{-0.5em}
\resizebox{0.99\textwidth}{!}{
\footnotesize
\begin{tabular}{c|l|ccccccccccccc|c|c}
\toprule
& Methods & Chess && Fire && Heads && Ofﬁce && Pumpkin && Kitchen && Stairs & Average & Mapping Time\\
\midrule
\multirow{2}{*}{\makecell[c]{SCR}} 
& DSAC* \cite{brachmann2020dsacstar} & 1.9/1.11 && 1.9/1.24 && 1.1/1.82 && 2.6/1.18 && 4.2/1.41 && 3.0/1.70 && 4.2/1.42 & 2.7/1.41 & Hours \\
& ACE \cite{brachmann2023ACE} & 1.9/0.7 && 1.9/0.9 && 0.9/0.6 && 2.7/0.8 && 4.2/1.1 && 4.2/1.3 && 3.9/1.1 & 2.8/ 0.93& \boldblack{5 Minutes} \\
\midrule
\multirow{11}{*}{\makecell[c]{APR}}
& PoseNet(PN)\cite{Kendall15}         & 32/8.12   && 47/14.4   && 29/12.0   && 48/7.68   && 47/8.42   && 59/8.64   && 47/13.8 & 44/10.4 & Hours \\
& PN Learn $\sigma^2$\cite{Kendall17}   & 14/4.50   && 27/11.8   && 18/12.1   && 20/5.77   && 25/4.82   && 24/5.52   && 37/10.6   & 24/7.87 & Hours \\
& geo. PN\cite{Kendall17}              & 13/4.48   && 27/11.3   && 17/13.0   && 19/5.55   && 26/4.75   && 23/5.35   && 35/12.4   & 23/8.12 & Hours \\
& LSTM PN\cite{Walch17}              & 24/5.77   && 34/11.9   && 21/13.7   && 30/8.08   && 33/7.00   && 37/8.83   && 40/13.7   & 31/9.85 & Hours \\
& Hourglass PN\cite{Melekhov17}         & 15/6.17   && 27/10.8  && 19/11.6   && 21/8.48   && 25/7.01    && 27/10.2   && 29/12.5   & 23/9.53 & Hours  \\
& BranchNet\cite{Wu17}            & 18/5.17   && 34/8.99   && 20/14.2   && 30/7.05   && 27/5.10   && 33/7.40   && 38/10.3   & 29/8.30 & Hours  \\
& MapNet\cite{Brahmbhatt18}               & 8/3.25   && 27/11.7   && 18/13.3   && 17/5.15   && 22/4.02   && 23/4.93   && 30/12.1   & 21/7.77 & Hours \\
& Direct-PN\cite{chen21}            & 10/3.52   && 27/8.66   && 17/13.1   && 16/5.96   && 19/3.85   && 22/5.13   && 32/10.6  & 20/7.26 & Days \\
& MS-Transformer\cite{Shavit21multiscene}       & 11/4.66   && 24/9.60  && 14/12.2  && 17/5.66   && 18/4.44   && 17/5.94   && 17/5.94   & 18/7.28 & Hours \\
& DFNet (VGG) \cite{chen2022dfnet}         & 5/1.88   && 17/6.45   && 6/3.63   && 8/2.48   && 10/2.78   && 22/5.45   && 16/3.29   & 12/3.71 & Days \\
& DFNet (EB0) \cite{chen2022dfnet}         & 3/1.15   && 9/3.71   && 8/6.08   && 7/2.14   && 10/2.76  && 9/2.87   && 11/5.58   & 8/3.47 & Days\\
& LENS \cite{Moreau21}         & 3/1.3   && 10/3.7   && 7/5.8   && 7/1.9   && 8/2.2  && 9/2.2   && 14/3.6   & 8/3.00 & Days\\
& $\marepo$ (Ours)     & 2.6/1.35   && 2.5/1.42   && 2.3/2.21   && 3.6/1.44 && 4.2/1.55  && 5.1/1.99   && 6.7/1.83   & 3.9/1.68 & \boldblack{5 Minutes} \\
& $\marepo_{S}$ (Ours)  & \boldblack{2.1/1.24} && \boldblack{2.3/1.39}   && \boldblack{1.8/2.03}   && \boldblack{2.8/1.26}   && \boldblack{3.5/1.48}   && \boldblack{4.2/1.71}  && \boldblack{5.6/1.67} & \boldblack{3.2/1.54}    & $\leq$ 15 Minutes\\

\bottomrule
\end{tabular}
}
\vspace{-0.5em}
\end{table*} 
\begin{table*}[t]
\footnotesize
\caption{\textbf{Pose accuracy comparison on the outdoor Wayspots dataset.}
Results are reported as the percentage of frames below $10cm/5\degree$ and $0.5m/5\degree$ pose error.
The map-relative pose regressors $\mathcal{M}$ of our $\marepo_{S}$ experiment are fine-tuned in $\approx 1$ minute for each scene.
}
\label{table:2}
\vspace{-0.5em}
\centering
\resizebox{\linewidth}{!}{
\footnotesize
\begin{tabular}{l|cc|cccc}
\toprule
Scene & DSAC* & ACE & PN & MST & $\marepo$ & $\marepo_{S}$   \\
& \cite{brachmann2020dsacstar} & \cite{brachmann2023ACE} & \cite{Kendall15,Kendall16,Kendall17} & \cite{Shavit21multiscene} & Ours & Ours  \\
\midrule
Throughput (fps)    &   17.9       &    17.9 &  \boldblack{166.7}      &      28.4         &     55.6     & 55.6        \\
\midrule
Bears        & 82.6\%/91.6\%  & 80.7\%/92.6\%  & 12.9\%/35.7\% & 0.5\%/12.8\%   & \boldblack{80.7\%}/99.3\%             & \boldblack{80.7\%}/\boldblack{99.5\%}   \\
Cubes        & 83.8\%/98.1\%  & 97.0\%/98.1\%  & 0.0\%/0.4\%    & 0.00\%/9.9\%  & \boldblack{72.4\%}/\boldblack{96.9\%} & 71.8\%/\boldblack{96.9\%}  \\
Inscription  & 54.1\%/69.7\%  & 49.0\%/69.6\%  & 1.1\%/6.3\%   & 1.3\%/9.7\%    & \boldblack{37.8\%}/\boldblack{74.2\%} & 37.1\%/74.1\%  \\
Lawn         & 34.7\%/38.0\%    & 35.8\%/38.5\%  & 0.0\%/0.2\%    & 0.0\%/0.0\%    & 32.6\%/\boldblack{41.6\%}          & \boldblack{34.2\%}/41.1\%   \\
Map          & 56.7\%/87.1\%  & 56.5\%/84.7\%  & 14.9\%/49.1\% & 5.6\%/25.7\%   & 53.9\%/87.7\%                         & \boldblack{55.1\%}/\boldblack{87.9\%}   \\
Square Bench & 69.5\%/97.9\%  & 66.7\%/97.8\%  & 0.0\%/3.0\%    & 0.0\%/0.0\%   & 68.6\%/\boldblack{100\%}              & \boldblack{70.7\%}/\boldblack{100\%}   \\
Statue       & 0.0\%/0.0\%     & 0.0\%/0.0\%     & 0.0\%/0.0\%     & 0.0\%/0.0\%    & 0.0\%/0.0\%                       & 0.0\%/0.0\%      \\
Tendrils     & 25.1\%/26.5\%  &34.9\%/36.8\%   & 0.0\%/0.0\%     & 0.9\%/23.6\% & 27.9\%/33.4\%                         & \boldblack{29.3\%}/\boldblack{34.8\%}   \\
The Rock     & 100\%/100\%     & 100\%/100\%     & 24.2\%/77.5\% & 10.7\%/52.6\%&98.1\%/\boldblack{100\%}               &\boldblack{99.8\%}/\boldblack{100\%}     \\
Winter Sign  & 0.2\%/5.7\%    & 1.0\%/7.6\%    & 0.0\%/0.0\%     & 0.0\%/0.0\%  & 0.0\%/\boldblack{0.7\%}               & 0.0\%/0.3\%     \\
\midrule
Average      & 50.7\%/61.5\%  & 52.2\%/62.6\%  & 5.3\%/17.2\%  & 1.9\%/13.4\%   & 47.2\%/63.4\%                         & \boldblack{47.9\%}/\boldblack{63.5\%}   \\
\bottomrule
\end{tabular}
}
\vspace{-1em}
\end{table*}

\subsection{Implementation Details}
\label{sec:implementation}
In this section, we provide details of the process used to train the \marepo network.
We first detail the generation of training data, then describe the architectural configuration.

\paragraph{Training Data Generation}
In our implementation, we employ the Accelerated Coordinate Encoding architecture and training protocol proposed in \cite{brachmann2023ACE} as $\mathcal{G}$ to train the scene-specific geometry prediction network $\mathcal{G_S}$ for each scene $S$ in the training dataset, since that allows the training of scene-specific coordinate-regression networks in a speedy fashion ($\sim$5 minutes for each new scene).
To train $\mathcal{M}$, we use 450 scenes (indexing from 0 to 459, excluding 200-209) from the Map-Free Dataset~\cite{arnold2022mapfree}.
Each image in the dataset has an associated ground truth camera pose computed by the authors of the dataset via SfM~\cite{colmap1}.
The data includes around 500K frames, with each scene containing images scanning a small outdoor location.
Frames from each scan have been split into mapping and query, with $\approx500$ mapping frames and $\approx500$ queries.
In practice, we train 900 scene-specific coordinate regressors $\mathcal{G_S}$ using frames from each scene and its two splits.
Given the efficient scaling capability of the method, we are able to generate a vast amount of 2D-3D correspondences between image pixels and scene coordinates by applying each $\mathcal{G_S}$ to the frames of the unseen split of its corresponding scan.
We use data augmentation during the generation of the correspondences, processing 16 variants of each frame.
Specifically, we apply random image rotations of up to $15\degree$; rescale each frame to $0.67 \sim 1.5$ times its original resolution; and, finally, extract random crops.
We save the scene coordinate maps output of the preprocessing, together with their corresponding masks indicating which pixels are valid after rotation, the augmented intrinsics, and camera poses.
This forms the fixed dataset we use to train the map-relative pose regressor $\mathcal{M}$.

Additionally, we perform online data augmentation by randomly jittering the scene coordinates by $\pm 1m$ and rotating them by up to $180\degree$ to further increase data diversity. Note that the random jittering is applied image-wise, i.e., all scene coordinates of an input frame are perturbed by the same transform.
We do this to avoid overfitting, ensuring that the network $\mathcal{M}$ does not learn an absolute pose for each frame but rather a pose relative to the scene coordinates.

\paragraph{Network Configuration}
The scene-specific networks $\mathcal{G_S}$ 
process images having the shortest side 480 pixels long and output dense scene coordinate maps with 8x smaller resolution.
The map-relative pose regressor $\mathcal{M}$ is built upon a cascade of linear-attention transformer blocks~\cite{sun2021loftr} with $d_{model}=256$ and $h=8$ parallel attention layers.
For the 3D position embedding, we prudently choose $m=5$ frequency bands due to presence of potentially noisy input.

\paragraph{Training  and Hardware Details}
We train $\mathcal{M}$ using 8 NVIDIA V100 GPUs with a batch size of 64.
We use the AdamW~\cite{loshchilov2017decoupled} optimizer with a learning rate between $3e^{-4}$ to $2e^{-3}$ with a 1-cycle scheduler~\cite{smith2019super}.
The model is trained for $\approx10$ days, iterating through the dataset for 150 epochs.

During inference our entire model, including the scene-specific network $\mathcal{G_S}$ and the map-relative pose regressor $\mathcal{M}$, requires only one GPU, and can estimate camera poses with a real-time throughput, as later shown in~\cref{table:2}.

\subsection{Quantitative Evaluation}
In the following paragraphs we show the performance of \marepo on two public datasets: one depicting indoor scenes, and one outdoor.
We show that the proposed map-relative pose regressor module $\mathcal{M}$ can generalize its predictions to previously unseen scenes, thanks to the scene-specific geometry prediction network $\mathcal{G_S}$ providing it with 2D-3D correspondences in each scene's metric space.

\vspace{-1em}
\paragraph{7-Scenes Dataset}
We first evaluate our method on the Microsoft 7-Scenes dataset~\cite{Glocker13,Shotton13}, an indoor relocalization dataset that provides up to 7000 mapping images per scene.
Each scene covers a limited area (between $1m^3$ and $18m^3$); despite that, previous APR methods require tens of hours or even several days~\cite{Moreau21} to train a model to relocalize in them.
This is nonideal in a practical scenario as the appearance of the scene might have changed within that time frame, thus rendering the trained APR out of date.
Conversely, $\marepo$ requires only minutes of training time ($\approx 5$) for each new scene to generate a geometry-prediction network $\mathcal{G_S}$ specifically tuned for the target environment.
We compare our method with prior Pose Regression approaches in~\cref{table:1}, showing that \marepo is not only a partly scene-agnostic approach that enjoys the fastest mapping time of all APR-based methods, but also obtains  $\approx 50\%$ better average performance (in terms of median error).
We also show the performance of a fine-tuned variant of our method, $\marepo_S$, where, in addition to training the scene-specific $\mathcal{G_S}$ scene coordinate predictor, we also use the mapping frames to run two epochs of fine-tuning on the $\mathcal{M}$ regressor (see~\cref{sec:arch-finetuning}).
The fine-tuned model $\marepo_S$ achieves further improvements in average performance, requiring only between $1.5 \sim 10$ extra minutes of training time, resulting in the only single-frame pose regression-based method able to achieve a similar level of accuracy as one of the current best 3D geometry-based methods, while being more efficient in terms of computational resources required.

\vspace{-1.25em}
\paragraph{Wayspots Dataset}
We further evaluate our method on the Wayspots dataset~\cite{brachmann2023ACE,arnold2022mapfree}, which depicts challenging outdoor scenes that even current geometry-based methods struggle with.
The dataset contains scans of 10 different areas with associated ground truth poses provided by a visual-inertial odometry system~\cite{arkit,arcore}.
In~\cref{table:2} we show a comparison of the performance of the proposed $\marepo$ (as well as the $\marepo_S$ models, fine-tuned on the mapping frames of each scene) with two APR-based approaches we reproduced; we also include a comparison with two scene coordinate regression approaches: DSAC*~\cite{brachmann2020dsacstar} and the current  state-of-the-art on Wayspots, ACE~\cite{brachmann2023ACE}.
\marepo significantly outperforms previous APR-based methods -- such as PoseNet~\cite{Kendall15} and MS-Transformers~\cite{Shavit21multiscene} -- that require on average several hours of training time, and compares favorably with geometry-based methods.
We show, for the first time, that an end-to-end image-to-pose regression method relying on geometric priors can achieve a similar level of performance as methods that require the deployment of a (slower) robust solver to estimate the camera pose from a set of potentially noisy 2D-3D correspondences.
More specifically, $\marepo$ requires only five minutes to train a network encoding the location of interest within the weights of the $\mathcal{G_S}$ scene-specific coordinate regressor and (\textit{optionally}) approximately one minute to fine-tune the map-relative regressor $\mathcal{M}$ (as the Wayspot scans have significantly less frames then the 7-Scenes scenes above).
At inference time \marepo (or its fine-tuned variant) can perform inference at $\approx 56$ frames per second, making it not only accurate, but also extremely efficient in comparison to other methods.

\begin{table}
\centering
\footnotesize
\begin{tabular}{lcc}
\toprule
\multirow{2}{*}{Model} & \multicolumn{2}{c}{Accuracy} \\
~ & $5cm/5\degree$ & $10cm/5\degree$ \\
\midrule
Full Architecture (\marepo)   & \textbf{16.6\%} &  \textbf{39.6\%} \\
- Re-Attention    & 10.9\% & 28.3\% \\
- Dynamic P.E.    &  3.9\% & 18.6\% \\
\bottomrule
\end{tabular}
\caption{
We gradually remove \textit{Re-Attention} and \textit{Dynamic Positional Encoding} and report the $\%$ of frames relocalized within $5cm/5\degree$ and $10cm/5\degree$.
}
\vspace{-1em}
\label{table:3 ablation}
\end{table}
\begin{table}
\centering
\footnotesize
\begin{tabular}{cccc}
\toprule
\multirow{2}{*}{\# T Blocks}  & \multirow{2}{*}{$d_{model}$} & \multicolumn{2}{c}{Accuracy} \\
~ & ~ & $5cm/5\degree$ & $10cm/5\degree$ \\
\midrule
4 & 128 & 5.8\% & 22.2\% \\
8 & 128 &  14.7\% & 39.1\%\\
12 & 128 & 16.6\% & 39.6\% \\
12 & 256 & \boldblack{19.0\%} & \boldblack{43.5\%}\\
\bottomrule
\end{tabular}
\caption{
Effect on performance of different dimensionality choices in the pose regressor's model.
\textit{\# T Blocks} denotes the number of transformer blocks used in the model.
$d_{model}$ denotes the width of the transformer layers. 
}
\vspace{-2em}
\label{table:4 ablation}
\end{table}

\subsection{Ablation Experiments}
In the following, we provide additional insights into the design choices adopted whilst designing our method.

\vspace{-1.25em}
\paragraph{Architecture Ablation}
We run several controlled experiments to justify our architectural design.
Note that, for the experiments in this subsection, we trained the transformer-based pose regressor for 50 epochs instead of 150 as in the main experiments.
This allows us to complete each experiment in approximately two days without affecting the relative ranking of results in the ablations.
For the first experiment, see~\cref{table:3 ablation}, we train a smaller transformer $\mathcal{M}$ (with $d_{model}=128$) and gradually remove the \textit{Re-Attention} and \textit{Dynamic Positional Encoding} components to evaluate their impact on the performance of the Wayspots dataset.
The pose accuracy is shown as the percentage of frames relocalized within $5cm/5\degree$ and $10cm/5\degree$ error.
The table shows consistent degradation without the proposed components.

Next, we show the impact of diverse model configurations by deploying different numbers of transformer blocks and $d_{model}$ dimensions in~\cref{table:4 ablation}.
We experimented with training even larger models with $d_{model}=512$ and 16/20 transformer blocks, but found they necessitated substantially more GPU resources and time.
We therefore cannot recommend them given the performance-time trade-offs.

\begin{table}
\centering
\footnotesize
\begin{tabular}{lcc}
\toprule
\multirow{2}{*}{Model} & \multicolumn{2}{c}{Accuracy} \\
~ & $10cm/5\degree$ & $50cm/5\degree$\\
\midrule
Per-scene \marepo & 0.7\%   & 6.0\% \\
Per-scene $\mathcal{M}$ & 2.9\%   & 18.7\% \\
\marepo (Ours) & \boldblack{47.2\%}   & \boldblack{63.4\%} \\
\bottomrule
\end{tabular}
\caption{
Effect of different training strategies.
Per-scene \marepo trains the entire network from scratch for every new mapping scene.
In per-scene $\mathcal{M}$, we train only the regressor from scratch, on top of a pre-trained $\mathcal{G_S}$.
Finally, \marepo is trained over the entire training dataset and can generalize well to unseen scenes.
}
\vspace{-2em}
\label{table: 5 ablation}
\end{table}

\vspace{-1.25em}
\paragraph{Per-Scene Training}
The proposed pose regressor component $\mathcal{M}$ is designed to be scene-agnostic and has been trained on a large corpus of data.
Still, we are interested in evaluating its performance when trained ad-hoc for individual scenes, similar to existing APR-based methods.
We conduct two experiments where, instead of using the full training set, we train scene-specific models using mapping sequences from the Wayspots dataset, as shown in~\cref{table: 5 ablation}.
First, the models are trained from scratch, i.e., both the scene geometry regressor $\mathcal{G_S}$ and the map-relative pose regressor $\mathcal{M}$ are trained as a single entity, similar to PoseNet~\cite{Kendall15}.
The results show extremely poor performance, likely due to the model's inability to learn explicit 3D geometry relations of the scene.
For the second variant, we assume a pre-trained $\mathcal{G_S}$ is provided, then train $\mathcal{M}$ from scratch for each scene.
We see results in a similar order of magnitude as other APR methods, such as PoseNet~\cite{Kendall15} or MST~\cite{Shavit21multiscene} (cf.~\cref{table:2}); still, this training approach performed quite poorly compared to the full \marepo model, where $\mathcal{M}$ is trained on a large-scale dataset to predict truly generic and scene-independent map-relative poses.
\section{Conclusion}
\label{sec:conclusion}

In conclusion, our paper introduces \marepo, a novel approach in Pose Regression that combines the strengths of a scene-agnostic pose regression network with a strong geometric prior provided by a fast-training scene-specific metric representation.
The method addresses the limitations of previous APR techniques, offering both scalability and precision in predicting accurate scale-metric poses across diverse scenes.
We demonstrate \marepo's superior accuracy and its capability for rapid adaptation to new scenes compared to existing APR methods on two datasets.
Additionally, we show how integrating the transformer-based network architecture with dynamic positional encoding ensures robustness to varying camera parameters, establishing \marepo as a versatile and efficient solution for regression-based visual relocalization.

\appendix
\clearpage
\maketitlesupplementary

\section{Supplementary Ablations}
\subsection{Impact of auxiliary losses}
In \cref{table:ablation6 auxiliary loss} we present an analysis of the impact of incorporating auxiliary losses $\mathcal{L}_{\hat{P_0}}, \mathcal{L}_{\hat{P_1}}$ into our model training protocol, contrasted with the model devoid of such losses.
As mentioned in Section 3.3 of the main paper, we found this implementation beneficial to the overall pose regression performance.
\begin{table}[h]
\centering
\footnotesize
\begin{tabular}{lcc}
\toprule
 \multirow{2}{*}{Architecture} & \multicolumn{2}{c}{Accuracy} \\
 ~ & $10cm/5\degree$ & $0.5m/5\degree$ \\
\midrule
\marepo w/ auxiliary losses (Ours) & \textbf{47.2\%} & \textbf{63.4\%} \\
\marepo w/o auxiliary losses &  45.7\% & 62.2\%\\
\bottomrule
\end{tabular}
\caption{
Performance of \marepo trained with and without auxiliary losses as in Equation 7 of the main paper.
}
\label{table:ablation6 auxiliary loss}
\end{table}

\subsection{Impact of rotation representation: 9D SVD orthogonalization  vs. 6D. Gram-Schmidt}
Additionally, we investigated the effects of utilizing alternative rotation representations on our model's performance.
For example, Levinson \textit{et al.}~\cite{levinson2020analysis} demonstrated that SVD orthogonalization facilitates a continuous mapping of a 9D representation onto SO(3), potentially improving pose prediction accuracy beyond that achievable with a 6D representation~\cite{Zhou_2019_CVPR} used in our model. 
We replaced our 6D with the 9D representation and trained the full \marepo models to assess the differences.
The findings indicate that, within our model's framework, the prediction accuracy for 9D rotations marginally lags behind that of 6D rotations (\cref{table:9D_vs_6D}), thereby verify our design choice in the paper.
\begin{table}[h]
\centering
\footnotesize
\begin{tabular}{lcc}
\toprule
 {Accuracy} & \marepo (9D) & \marepo (6D) \\
\midrule
$10cm/5\degree$  & 46.8\% &\textbf{ 47.2\%} \\
\bottomrule
\end{tabular}
\caption{
Ablation on rotation representations using the \marepo model. Accuracies are reported on Wayspots.
}
\label{table:9D_vs_6D}
\end{table}

\subsection{Impact of the SCR component}
We use two methods to study the impact of the choice of Scene Coordinate Regression component on the pose estimation performance.
First, we replaced the pretrained ACE backbone with a VGG network, then retrained the scene-specific SCRs.
The SCRs' outputs were then passed to our pretrained pose regressor $\mathcal{M}$.
As indicated in \cref{table:rebuttal3}, the choice of SCR does indeed affect the pose regressor's accuracy.
However, \marepo also displays robustness to the quality of the input scene coordinates, as the overall performance degradation is not large, demonstrating the capability of our approach to predict accurate poses from scene coordinates generated by different means.

Furthermore, we performed quantitative experiments adding random noise to the scene coordinates passed to $\mathcal{M}$.
Specifically, we applied randomly generated noise of different magnitudes (up to 10cm, and up to 50cm) to a variable proportion of the scene coordinates.
We show that \marepo is able to cope with large proportions of errors in the input coordinates, without significant drops in performance (up to 60\% of the coordinates can be perturbed with 10cm noise, and up to 40\% for 50cm noise) (see~\cref{table:rebuttal-noise}).
\begin{table}[h]

\centering
\footnotesize
\begin{tabular}{lcc}
\toprule
 {Accuracy} & ACE backbone SCR + $\mathcal{M}$ & VGG backbone SCR + $\mathcal{M}$ \\
\midrule
$10cm/5\degree$ & 47.2\% & 46.0\%\\
\bottomrule
\end{tabular}
\caption{
Effect of different scene coordinate regression backbones on the accuracy of the downstream regressor $\mathcal{M}$ on Wayspots.
}
\label{table:rebuttal3}

\end{table}

\begin{table}[h]

\centering
\footnotesize
\begin{tabular}{lccccccc}
\toprule
 {SCR Noise} & $0\%$ & $20\%$ & $40\%$ & $60\%$ & $80\%$ & $100\%$ \\
\midrule
$10cm$ & 47.2\% & 46.9 & 46.0 & 44.5 & 38.2 & 26.9\\
$50cm$ & 47.2\% & 46.3 & 43.2 & 21.1 & 10.3 & 0.3\\
\bottomrule
\end{tabular}
\caption{
Effect of increasing amounts of random noise applied to the SCR predictions.
The top row indicates the proportion of the pixels in each scene coordinate map affected by uniform noise with maximum value indicated at the beginning of each row.
We report the $10cm/5\degree$ accuracy on Wayspots.
}
\label{table:rebuttal-noise}

\end{table}

\section{Supplementary Video}
To complement our quantitative analysis, we provide a supplementary video offering a qualitative perspective, primarily focusing on visually comparing the predicted camera trajectories.
The trajectories are superimposed on point clouds rendered from the respective scenes, providing an intuitive understanding of each method's performance.

The first segment of the video showcases a comparative analysis of our approach against other open-sourced APR-based methods on the 7-Scenes dataset~\cite{Glocker13,Shotton13}, where we compare vs. PoseNet\cite{Kendall15,Kendall17} and DFNet(EB0)\cite{chen2022dfnet}; and the Wayspots dataset~\cite{brachmann2023ACE,arnold2022mapfree}, where we compare vs. PoseNet and MS-Transformer\cite{Shavit21multiscene}, note that both train their models in under one day.
For PoseNet, we have utilized the PyTorch implementation provided by Chen~\textit{et al.}\cite{chen21}. 

Moreover, in the second segment of the video, we show that \marepo compares well qualitatively with the Accelerated Coordinate Encoding\cite{brachmann2023ACE} structure-based method.
This comparison demonstrates that our method achieves similar accuracy to ACE, with the added benefit of producing smoother trajectory estimations in certain scenarios.
Notably, our approach provides a faster throughput during inference, underscoring its practical applicability in demanding scenarios.

\section{Experiments on 12-Scenes dataset}
We show experimental results on the 12-Scenes dataset~\cite{valentin2016learning}
in \cref{table:12_scenes}.
We compare \marepo to the baseline APR methods PoseNet and MS-Transformer.
Since the original PoseNet code was implemented on Caffe, we used the open-sourced code from \cite{chen21}.
The results show that \marepo significantly outperforms the baseline APRs, which is consistent with the behavior shown in the main paper compared to the benchmark APR approaches.
\begin{table}[h]
\centering
\footnotesize
\begin{tabular}{lccc}
\toprule
 Scene & PoseNet & MST & \marepo \\
\midrule
Apt.1 Kitchen       & 14.3\%    & 3.4\%    & 98.0\%\\
Apt.1 Living        & 11.2\%    & 9.7\%    & 98.6\%\\
\midrule
Apt.2 Bed           & 18.1\%    & 2.9\%    & 96.0\%\\
Apt.2 Kitchen       & 38.6\%    & 13.8\%   & 100\%\\
Apt.2 Living        & 13.5\%    & 4.6\%    & 99.7\%\\
Apt.2 Luke          & 9.1\%     & 4.8\%    & 89.4\%\\
\midrule
Office 1 Gates 362  & 34.5\%    & 14.0\%   & 97.2\%\\
Office 1 Gates 381  & 8.1\%     & 4.1\%    & 84.6\%\\
Office 1 Lounge     & 17.1\%    & 14.1\%   & 93.9\%\\
Office 1 Manolis    & 13.7\%    & 8.9\%    & 94.8\%\\
\midrule
Office 2 Floor 5a   & 5.2\%     & 1.4\%    & 90.5\%\\
Office 2 Floor 5b   & 5.2\%     & 7.2\%    & 83.5\%\\
\midrule
Average  & 13.5\% & 7.4\% &\textbf{93.9\%}  \\
\midrule
median error   & 9.4cm/3.9$\degree$ & 11.1cm/5.5$\degree$ & \textbf{2.6cm/1.3$\degree$}  \\
\bottomrule
\end{tabular}
\caption{
Performance on the 12-Scenes \cite{valentin2016learning} dataset. The accuracy is reported as percentage of query frames localized within $5cm/5\degree$.
}
\label{table:12_scenes}
\end{table}

{
    \small
    \bibliographystyle{ieeenat_fullname}
    \bibliography{main}

\begin{thebibliography}{59}
\providecommand{\natexlab}[1]{#1}
\providecommand{\url}[1]{\texttt{#1}}
\expandafter\ifx\csname urlstyle\endcsname\relax
  \providecommand{\doi}[1]{doi: #1}\else
  \providecommand{\doi}{doi: \begingroup \urlstyle{rm}\Url}\fi

\bibitem[Apple()]{arkit}
Apple.
\newblock \href{https://developer.apple.com/documentation/arkit/configuration_objects/understanding_world_tracking}{ARKit}.
\newblock Accessed: 26 March 2024.

\bibitem[Arnold et~al.(2022)Arnold, Wynn, Vicente, Garcia-Hernando, Monszpart, Prisacariu, Turmukhambetov, and Brachmann]{arnold2022mapfree}
Eduardo Arnold, Jamie Wynn, Sara Vicente, Guillermo Garcia-Hernando, {\'{A}}ron Monszpart, Victor~Adrian Prisacariu, Daniyar Turmukhambetov, and Eric Brachmann.
\newblock Map-free visual relocalization: Metric pose relative to a single image.
\newblock In \emph{ECCV}, 2022.

\bibitem[Brachmann and Rother(2019)]{brachmann2019ngransac}
Eric Brachmann and Carsten Rother.
\newblock {N}eural- {G}uided {RANSAC}: {L}earning where to sample model hypotheses.
\newblock In \emph{ICCV}, 2019.

\bibitem[Brachmann and Rother(2021)]{brachmann2020dsacstar}
Eric Brachmann and Carsten Rother.
\newblock Visual camera re-localization from {RGB} and {RGB-D} images using {DSAC}.
\newblock \emph{IEEE TPAMI}, 2021.

\bibitem[Brachmann et~al.(2017)Brachmann, Krull, Nowozin, Shotton, Michel, Gumhold, and Rother]{Brachmann17}
Eric Brachmann, Alexander Krull, Sebastian Nowozin, Jamie Shotton, Frank Michel, Stefan Gumhold, and Carsten Rother.
\newblock {DSAC - Differentiable RANSAC for Camera Localization}.
\newblock In \emph{CVPR}, 2017.

\bibitem[Brachmann et~al.(2023)Brachmann, Cavallari, and Prisacariu]{brachmann2023ACE}
Eric Brachmann, Tommaso Cavallari, and Victor~Adrian Prisacariu.
\newblock Accelerated coordinate encoding: Learning to relocalize in minutes using rgb and poses.
\newblock In \emph{CVPR}, 2023.

\bibitem[Brahmbhatt et~al.(2018)Brahmbhatt, Gu, Kim, Hays, and Kautz]{Brahmbhatt18}
S. Brahmbhatt, J. Gu, K. Kim, J. Hays, and J. Kautz.
\newblock {Geometry-Aware Learning of Maps for Camera Localization}.
\newblock In \emph{CVPR}, 2018.

\bibitem[Camposeco et~al.(2019)Camposeco, Cohen, Pollefeys, and Sattler]{compression2019cvpr}
Federico Camposeco, Andrea Cohen, Marc Pollefeys, and Torsten Sattler.
\newblock Hybrid scene compression for visual localization.
\newblock In \emph{CVPR}, 2019.

\bibitem[Carion et~al.(2020)Carion, Massa, Synnaeve, Usunier, Kirillov, and Zagoruyko]{carion2020DeTR}
Nicolas Carion, Francisco Massa, Gabriel Synnaeve, Nicolas Usunier, Alexander Kirillov, and Sergey Zagoruyko.
\newblock End-to-end object detection with transformers.
\newblock In \emph{ECCV}, 2020.

\bibitem[Chen et~al.(2021{\natexlab{a}})Chen, Snavely, and Makadia]{chen2021wide}
Kefan Chen, Noah Snavely, and Ameesh Makadia.
\newblock Wide-baseline relative camera pose estimation with directional learning.
\newblock In \emph{CVPR}, 2021{\natexlab{a}}.

\bibitem[Chen et~al.(2021{\natexlab{b}})Chen, Wang, and Prisacariu]{chen21}
Shuai Chen, Zirui Wang, and Victor Prisacariu.
\newblock {Direct-PoseNet}: Absolute pose regression with photometric consistency.
\newblock In \emph{3DV}, 2021{\natexlab{b}}.

\bibitem[Chen et~al.(2022)Chen, Li, Wang, and Prisacariu]{chen2022dfnet}
Shuai Chen, Xinghui Li, Zirui Wang, and Victor Prisacariu.
\newblock {DFN}et: {E}nhance absolute pose regression with direct feature matching.
\newblock In \emph{ECCV}, 2022.

\bibitem[Clark et~al.(2017)Clark, Wang, Markham, Trigoni, and Wen]{clark2017vidloc}
Ronald Clark, Sen Wang, Andrew Markham, Niki Trigoni, and Hongkai Wen.
\newblock Vidloc: A deep spatio-temporal model for 6-dof video-clip relocalization.
\newblock In \emph{CVPR}, 2017.

\bibitem[Dong et~al.(2022)Dong, Wang, Zhuang, Kannala, Pollefeys, and Chen]{dong2022visual}
Siyan Dong, Shuzhe Wang, Yixin Zhuang, Juho Kannala, Marc Pollefeys, and Baoquan Chen.
\newblock Visual localization via few-shot scene region classification.
\newblock In \emph{3DV}, 2022.

\bibitem[Dosovitskiy et~al.(2020)Dosovitskiy, Beyer, Kolesnikov, Weissenborn, Zhai, Unterthiner, Dehghani, Minderer, Heigold, Gelly, et~al.]{dosovitskiy2020image}
Alexey Dosovitskiy, Lucas Beyer, Alexander Kolesnikov, Dirk Weissenborn, Xiaohua Zhai, Thomas Unterthiner, Mostafa Dehghani, Matthias Minderer, Georg Heigold, Sylvain Gelly, et~al.
\newblock An image is worth 16x16 words: Transformers for image recognition at scale.
\newblock \emph{arXiv preprint arXiv:2010.11929}, 2020.

\bibitem[En et~al.(2018)En, Lechervy, and Jurie]{en2018rpnet}
Sovann En, Alexis Lechervy, and Fr{\'e}d{\'e}ric Jurie.
\newblock Rpnet: An end-to-end network for relative camera pose estimation.
\newblock In \emph{ECCVW}, 2018.

\bibitem[Fischler and Bolles(1981)]{Fischler81}
Martin~A. Fischler and Robert~C. Bolles.
\newblock {Random sample consensus: a paradigm for model fitting with applications to image analysis and automated cartography}.
\newblock In \emph{CACM}, 1981.

\bibitem[Gao et~al.(2003)Gao, Hou, Tang, and Cheng]{gao2003complete}
Xiao-Shan Gao, Xiao-Rong Hou, Jianliang Tang, and Hang-Fei Cheng.
\newblock Complete solution classification for the perspective-three-point problem.
\newblock \emph{IEEE TPAMI}, 2003.

\bibitem[Glocker et~al.(2013)Glocker, Izadi, Shotton, and Criminisi]{Glocker13}
Ben Glocker, Shahram Izadi, Jamie Shotton, and Antonio Criminisi.
\newblock Real-time rgb-d camera relocalization.
\newblock In \emph{ISMAR}, 2013.

\bibitem[Google()]{arcore}
Google.
\newblock \href{https://developers.google.com/ar/develop/fundamentals}{ARCore}.
\newblock Accessed: 26 March 2024.

\bibitem[Humenberger et~al.(2020)Humenberger, Cabon, Guerin, Morat, Revaud, Rerole, Pion, de~Souza, Leroy, and Csurka]{kapture2020}
Martin Humenberger, Yohann Cabon, Nicolas Guerin, Julien Morat, Jérôme Revaud, Philippe Rerole, Noé Pion, Cesar de Souza, Vincent Leroy, and Gabriela Csurka.
\newblock Robust image retrieval-based visual localization using {Kapture}, 2020.

\bibitem[Katharopoulos et~al.(2020)Katharopoulos, Vyas, Pappas, and Fleuret]{katharopoulos2020transformers}
Angelos Katharopoulos, Apoorv Vyas, Nikolaos Pappas, and Fran{\c{c}}ois Fleuret.
\newblock Transformers are rnns: Fast autoregressive transformers with linear attention.
\newblock In \emph{ICML}, 2020.

\bibitem[Kendall and Cipolla(2016)]{Kendall16}
A. Kendall and R. Cipolla.
\newblock Modelling uncertainty in deep learning for camera relocalization.
\newblock In \emph{ICRA}, 2016.

\bibitem[Kendall and Cipolla(2017)]{Kendall17}
A. Kendall and R. Cipolla.
\newblock Geometric loss functions for camera pose regression with deep learning.
\newblock In \emph{CVPR}, 2017.

\bibitem[Kendall et~al.(2015)Kendall, Grimes, and Cipolla]{Kendall15}
A. Kendall, M. Grimes, and R. Cipolla.
\newblock Posenet: A convolutional network for real-time 6-dof camera relocalization.
\newblock In \emph{ICCV}, 2015.

\bibitem[Levinson et~al.(2020)Levinson, Esteves, Chen, Snavely, Kanazawa, Rostamizadeh, and Makadia]{levinson2020analysis}
Jake Levinson, Carlos Esteves, Kefan Chen, Noah Snavely, Angjoo Kanazawa, Afshin Rostamizadeh, and Ameesh Makadia.
\newblock An analysis of svd for deep rotation estimation.
\newblock \emph{NeurIPS}, 2020.

\bibitem[Li et~al.(2020)Li, Wang, Zhao, Verbeek, and Kannala]{li2020hierarchical}
Xiaotian Li, Shuzhe Wang, Yi Zhao, Jakob Verbeek, and Juho Kannala.
\newblock Hierarchical scene coordinate classification and regression for visual localization.
\newblock In \emph{CVPR}, pages 11983--11992, 2020.

\bibitem[Loshchilov and Hutter(2017)]{loshchilov2017decoupled}
Ilya Loshchilov and Frank Hutter.
\newblock Decoupled weight decay regularization.
\newblock \emph{ICLR}, 2017.

\bibitem[Lynen et~al.(2019)Lynen, Zeisl, Aiger, Bosse, Hesch, Pollefeys, Siegwart, and Sattler]{Lynen2019largescale}
Simon Lynen, Bernhard Zeisl, Dror Aiger, Michael Bosse, Joel Hesch, Marc Pollefeys, Roland Siegwart, and Torsten Sattler.
\newblock Large-scale, real-time visual-inertial localization revisited.
\newblock \emph{International Journal of Robotics Research}, 39(9), 2019.

\bibitem[Melekhov et~al.(2017)Melekhov, Ylioinas, Kannala, and Rahtu]{Melekhov17}
I. Melekhov, J. Ylioinas, J. Kannala, and E. Rahtu.
\newblock Image-based localization using hourglass networks.
\newblock In \emph{ICCVW}, 2017.

\bibitem[Mildenhall et~al.(2020)Mildenhall, Srinivasan, Tancik, Barron, Ramamoorthi, and Ng]{Mildenhall20nerf}
Ben Mildenhall, Pratul~P. Srinivasan, Matthew Tancik, Jonathan~T. Barron, Ravi Ramamoorthi, and Ren Ng.
\newblock {NeRF Representing scenes as neural radiance fields for view synthesis}.
\newblock In \emph{ECCV}, 2020.

\bibitem[Moreau et~al.(2021)Moreau, Piasco, Tsishkou, Stanciulescu, and de~La~Fortelle]{Moreau21}
Arthur Moreau, Nathan Piasco, Dzmitry Tsishkou, Bogdan Stanciulescu, and Arnaud de La~Fortelle.
\newblock {LENS}: Localization enhanced by nerf synthesis.
\newblock In \emph{CoRL}, 2021.

\bibitem[Moreau et~al.(2022)Moreau, Piasco, Tsishkou, Stanciulescu, and de~La~Fortelle]{moreau2022coordinet}
Arthur Moreau, Nathan Piasco, Dzmitry Tsishkou, Bogdan Stanciulescu, and Arnaud de La~Fortelle.
\newblock Coordinet: uncertainty-aware pose regressor for reliable vehicle localization.
\newblock In \emph{WACV}, 2022.

\bibitem[Naseer and Burgard(2017)]{Naseer17}
T. Naseer and W. Burgard.
\newblock Deep regression for monocular camera-based 6-dof global localization in outdoor environments.
\newblock In \emph{IROS}, 2017.

\bibitem[Panek et~al.(2022)Panek, Kukelova, and Sattler]{panek2022meshloc}
Vojtech Panek, Zuzana Kukelova, and Torsten Sattler.
\newblock Meshloc: Mesh-based visual localization.
\newblock In \emph{ECCV}, pages 589--609. Springer, 2022.

\bibitem[Purkait et~al.(2018)Purkait, Zhao, and Zach]{Purkait18}
Pulak Purkait, Cheng Zhao, and Christopher Zach.
\newblock Synthetic view generation for absolute pose regression and image synthesis.
\newblock In \emph{BMVC}, 2018.

\bibitem[Radwan et~al.(2018)Radwan, Valada, and Burgard]{Radwan18}
N. Radwan, A. Valada, and W. Burgard.
\newblock Vlocnet++: Deep multitask learning for semantic visual localization and odometry.
\newblock In \emph{IEEE Robotics and Automation Letters}, 2018.

\bibitem[Sarlin et~al.(2019)Sarlin, Cadena, Siegwart, and Dymczyk]{sarlin2019HFNet}
Paul-Edouard Sarlin, Cesar Cadena, Roland Siegwart, and Marcin Dymczyk.
\newblock From coarse to fine: Robust hierarchical localization at large scale.
\newblock In \emph{CVPR}, 2019.

\bibitem[Sarlin et~al.(2020)Sarlin, DeTone, Malisiewicz, and Rabinovich]{Sarlin20}
Paul-Edouard Sarlin, Daniel DeTone, Tomasz Malisiewicz, and Andrew Rabinovich.
\newblock Superglue: Learning feature matching with graph neural networks.
\newblock In \emph{CVPR}, 2020.

\bibitem[Sarlin et~al.(2021)Sarlin, Unagar, Larsson, Germain, Toft, Larsson, Pollefeys, Lepetit, Hammarstrand, Kahl, and Sattler]{sarlin21pixloc}
Paul-Edouard Sarlin, Ajaykumar Unagar, Måns Larsson, Hugo Germain, Carl Toft, Viktor Larsson, Marc Pollefeys, Vincent Lepetit, Lars Hammarstrand, Fredrik Kahl, and Torsten Sattler.
\newblock {Back to the Feature}: Learning robust camera localization from pixels to pose.
\newblock In \emph{CVPR}, 2021.

\bibitem[Sattler et~al.(2012)Sattler, Leibe, and Kobbelt]{Sattler12}
T. Sattler, B. Leibe, and L. Kobbelt.
\newblock Improving image-based localization by active correspondence search.
\newblock In \emph{ECCV}, 2012.

\bibitem[Sattler et~al.(2017)Sattler, Leibe, and Kobbelt]{Sattler17}
T. Sattler, B. Leibe, and L. Kobbelt.
\newblock {Efficient \& Effective Prioritized Matching for Large-Scale Image-Based Localization}.
\newblock In \emph{IEEE TPAMI}, 2017.

\bibitem[Sattler et~al.(2019)Sattler, Zhou, Pollefeys, and Leal-Taixe]{Sattler19}
T. Sattler, Q. Zhou, M. Pollefeys, and L. Leal-Taixe.
\newblock Understanding the limitations of cnn-based absolute camera pose regression.
\newblock In \emph{CVPR}, 2019.

\bibitem[Sch\"{o}nberger and Frahm(2016)]{colmap1}
Johannes~Lutz Sch\"{o}nberger and Jan-Michael Frahm.
\newblock Structure-from-motion revisited.
\newblock In \emph{CVPR}, 2016.

\bibitem[Shavit and Keller(2022)]{shavit2022camera}
Yoli Shavit and Yosi Keller.
\newblock Camera pose auto-encoders for improving pose regression.
\newblock In \emph{ECCV}, 2022.

\bibitem[Shavit et~al.(2021{\natexlab{a}})Shavit, Ferens, and Keller]{Shavit21}
Yoli Shavit, Ron Ferens, and Yosi Keller.
\newblock Paying attention to activation maps in camera pose regression.
\newblock In \emph{arXiv preprint arXiv:2103.11477}, 2021{\natexlab{a}}.

\bibitem[Shavit et~al.(2021{\natexlab{b}})Shavit, Ferens, and Keller]{Shavit21multiscene}
Yoli Shavit, Ron Ferens, and Yosi Keller.
\newblock Learning multi-scene absolute pose regression with transformers.
\newblock In \emph{ICCV}, 2021{\natexlab{b}}.

\bibitem[Shotton et~al.(2013)Shotton, Glocker, Zach, Izadi, Criminisi, and Fitzgibbon]{Shotton13}
Jamie Shotton, Ben Glocker, Christopher Zach, Shahram Izadi, Antonio Criminisi, and Andrew Fitzgibbon.
\newblock Scene coordinate regression forests for camera relocalization in rgb-d images.
\newblock In \emph{CVPR}, 2013.

\bibitem[Smith and Topin(2019)]{smith2019super}
Leslie~N Smith and Nicholay Topin.
\newblock Super-convergence: Very fast training of neural networks using large learning rates.
\newblock In \emph{Artificial intelligence and machine learning for multi-domain operations applications}, 2019.

\bibitem[Sun et~al.(2021)Sun, Shen, Wang, Bao, and Zhou]{sun2021loftr}
Jiaming Sun, Zehong Shen, Yuang Wang, Hujun Bao, and Xiaowei Zhou.
\newblock {LoFTR}: Detector-free local feature matching with transformers.
\newblock \emph{CVPR}, 2021.

\bibitem[T{\"{u}}rko\u{g}lu et~al.(2021)T{\"{u}}rko\u{g}lu, Brachmann, Schindler, Brostow, and Monszpart]{turkoglu2021visual}
Mehmet~{\"{O}}zg{\"{u}}r T{\"{u}}rko\u{g}lu, Eric Brachmann, Konrad Schindler, Gabriel Brostow, and \'{A}ron Monszpart.
\newblock {Visual Camera Re-Localization Using Graph Neural Networks and Relative Pose Supervision}.
\newblock In \emph{3DV}, 2021.

\bibitem[Valentin et~al.(2016)Valentin, Dai, Nie{\ss}ner, Kohli, Torr, Izadi, and Keskin]{valentin2016learning}
Julien Valentin, Angela Dai, Matthias Nie{\ss}ner, Pushmeet Kohli, Philip Torr, Shahram Izadi, and Cem Keskin.
\newblock Learning to navigate the energy landscape.
\newblock In \emph{3DV}. IEEE, 2016.

\bibitem[Walch et~al.(2017)Walch, Hazirbas, Leal-Taixe, Sattler, Hilsenbeck, and Cremers]{Walch17}
F. Walch, C. Hazirbas, L. Leal-Taixe, T. Sattler, S. Hilsenbeck, and D. Cremers.
\newblock Image-based localization using lstms for structured feature correlation.
\newblock In \emph{ICCV}, 2017.

\bibitem[Wang et~al.(2020)Wang, Chen, Lu, Zhao, Trigoni, and Markham]{atloc}
Bing Wang, Changhao Chen, Chris~Xiaoxuan Lu, Peijun Zhao, Niki Trigoni, and Andrew Markham.
\newblock Atloc: Attention guided camera localization.
\newblock In \emph{AAAI}, 2020.

\bibitem[Winkelbauer et~al.(2021)Winkelbauer, Denninger, and Triebel]{WinkelbauerICRA21}
Dominik Winkelbauer, Maximilian Denninger, and Rudolph Triebel.
\newblock Learning to localize in new environments from synthetic training data.
\newblock In \emph{ICRA}, 2021.

\bibitem[Wu et~al.(2017)Wu, Ma, and Hu]{Wu17}
J. Wu, L. Ma, and X. Hu.
\newblock {Delving Deeper into Convolutional Neural Networks for Camera Relocalization}.
\newblock In \emph{ICRA}, 2017.

\bibitem[Yang et~al.(2019)Yang, Bai, Tang, Li, Furukawa, and Tan]{yang2019sanet}
Luwei Yang, Ziqian Bai, Chengzhou Tang, Honghua Li, Yasutaka Furukawa, and Ping Tan.
\newblock {SANet}: {S}cene agnostic network for camera localization.
\newblock In \emph{ICCV}, 2019.

\bibitem[Zhou et~al.(2022)Zhou, Agostinho, O{\v{s}}ep, and Leal-Taix{\'e}]{zhou2022gomatch}
Qunjie Zhou, S{\'e}rgio Agostinho, Aljo{\v{s}}a O{\v{s}}ep, and Laura Leal-Taix{\'e}.
\newblock Is geometry enough for matching in visual localization?
\newblock In \emph{ECCV}, 2022.

\bibitem[Zhou et~al.(2019)Zhou, Barnes, Jingwan, Jimei, and Hao]{Zhou_2019_CVPR}
Yi Zhou, Connelly Barnes, Lu Jingwan, Yang Jimei, and Li Hao.
\newblock On the continuity of rotation representations in neural networks.
\newblock In \emph{CVPR}, 2019.

\end{thebibliography}
}



\end{document}